\documentclass{article}
\usepackage{spconf}
\usepackage{amsmath,amssymb,booktabs,graphicx}
\usepackage[hidelinks]{hyperref}

\title{Role-Aware Server Optimization for Multimodal Federated Learning
with Missing Modalities}

\name{Liuyin Shi, Wenbo He}
\address{McMaster University}

\begin{document}
\ninept
\maketitle

\begin{abstract}
Multimodal federated learning (MFL) trains a shared model by
aggregating client updates without exchanging raw data. Under
missing modalities, clients adapt active modality encoders and a shared
prediction block (fusion and classifier) to different input subsets.
This raises two server-side questions: which clients should
contribute to each component, and how should their aggregates be
incorporated across rounds? We observe that an absent-modality
encoder can still adapt through missing-input handling.
Availability-based aggregation can therefore omit its update while
retaining jointly learned prediction-block updates. Contributor
selection also shapes server momentum, a weighted history of
aggregated updates. We observe that prediction-block momentum can
improve or reduce performance depending on the encoder update policy.
We therefore propose Role-Aware Server Optimization (RASO), which
coordinates contributor selection with role-specific server updates.
RASO selects contributors through prediction dependence and inclusion
in local optimization, retains encoder momentum, and synchronizes
the shared prediction block to its current aggregate. Experiments
across datasets and fusion architectures demonstrate the effectiveness
of RASO under missing modalities.
\end{abstract}


\section{Introduction}
\label{sec:introduction}

Multimodal federated learning (MFL) enables distributed clients to train
a shared model without exchanging raw data
~\cite{mcmahan2017fedavg,feng2023fedmultimodal}. Missing modalities restrict
local observations and can bias learning toward available inputs
~\cite{fan2024overcome,yuan2026communication}. Prototype and representation
methods address learning from these incomplete observations
~\cite{bao2026pmcmfl,le2025mfcpl,nguyen2024fedmac,che2024leveraging,nguyen2026learning}.
The server must then combine updates to the active modality encoders and
shared prediction block (fusion and classifier), which are trained jointly
under each client's input conditions.

\begin{figure}[!t]
\centering
\includegraphics[width=\columnwidth]{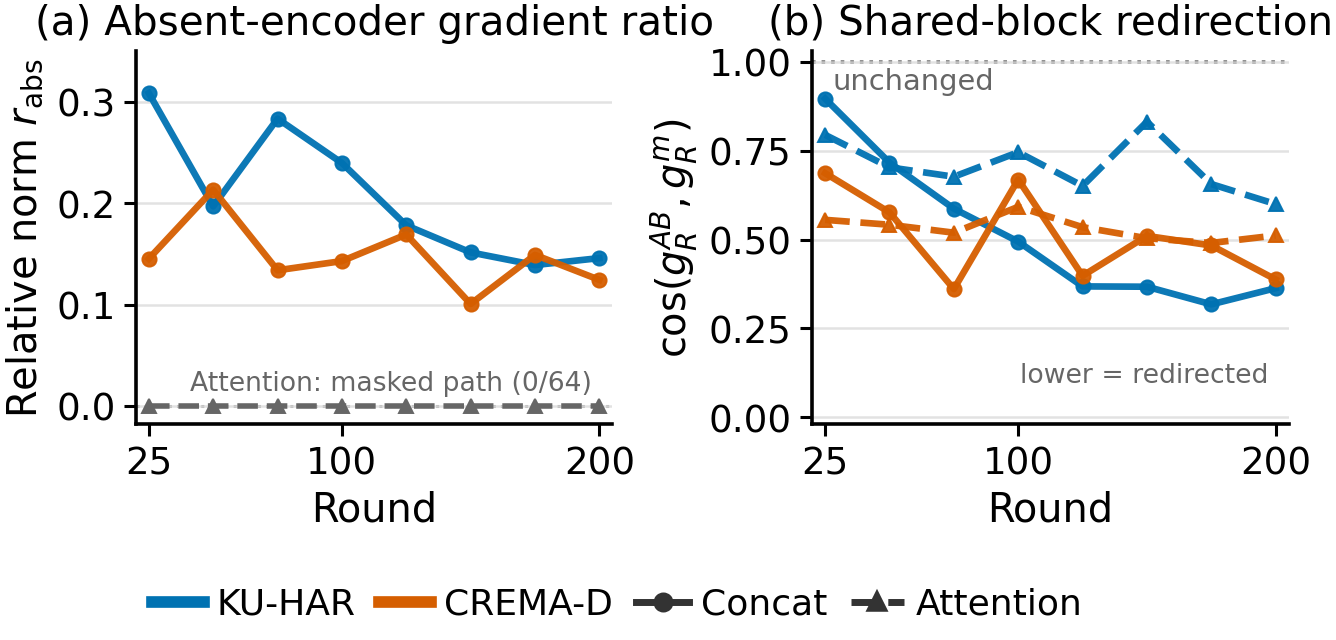}
\caption{
Paired complete and single-missing cross-entropy (CE) gradients on
identical examples and parameters. (a) Absent-encoder share $r_{\rm abs}$.
(b) Prediction-block cosine $\cos(g_R^{AB},g_R^m)$: $AB$ denotes
complete input, $m$ the retained modality. One indicates the same
direction; lower values indicate greater redirection.
}
\label{fig:gradient_observation}
\end{figure}

Modality selection~\cite{fan2024overcome,yuan2024fedmfs} and contribution
weighting~\cite{liu2025fedmobile} prioritize inputs and updates, while
block-wise sharing~\cite{chen2022fedmsplit,pranav2026blossom} specifies
exchanged components. Selecting a component's contributors also requires
distinguishing the modalities a client observes from the components involved
in its local learning (Fig.~\ref{fig:gradient_observation}). Under concatenation, an absent encoder can receive
task gradients through its zero-input response.
Availability-based aggregation can then omit its update while retaining
prediction-block updates learned alongside it.

These aggregation choices also determine which updates are retained across
rounds. FedOpt~\cite{reddi2021fedopt} applies server optimization to client
aggregates; momentum retains a weighted history of those updates.
Parameter-specific sharing and optimization
~\cite{alam2022fedrolex,10.1007/978-981-92-3520-9_47}, including FedPVR's
classification-layer variance reduction~\cite{li2023fedpvr}, allow different
treatments for model blocks. For encoders and the prediction block, the
pairing of their update policies matters: encoder updates change the
features used for prediction, while prediction-block updates change encoder
gradients. We observe that prediction-block momentum improves performance
with current encoder aggregation but reduces it with persistent encoder
updates (Table~\ref{tab:raso_lifecycle}).

We therefore propose \emph{Role-Aware Server Optimization} (RASO), with
the following contributions:
(i) We characterize how missing inputs affect component participation
and gradient directions, and identify interactions between encoder and
prediction-block server-update policies.
(ii) We develop a server-update scheme that selects contributors through
prediction dependence and local optimization, retains encoder momentum,
and synchronizes the prediction block to its current aggregate.

\section{Role-Aware Server Optimization}
\label{sec:method}

\begin{figure*}[!t]
\centering
\includegraphics[width=\textwidth]{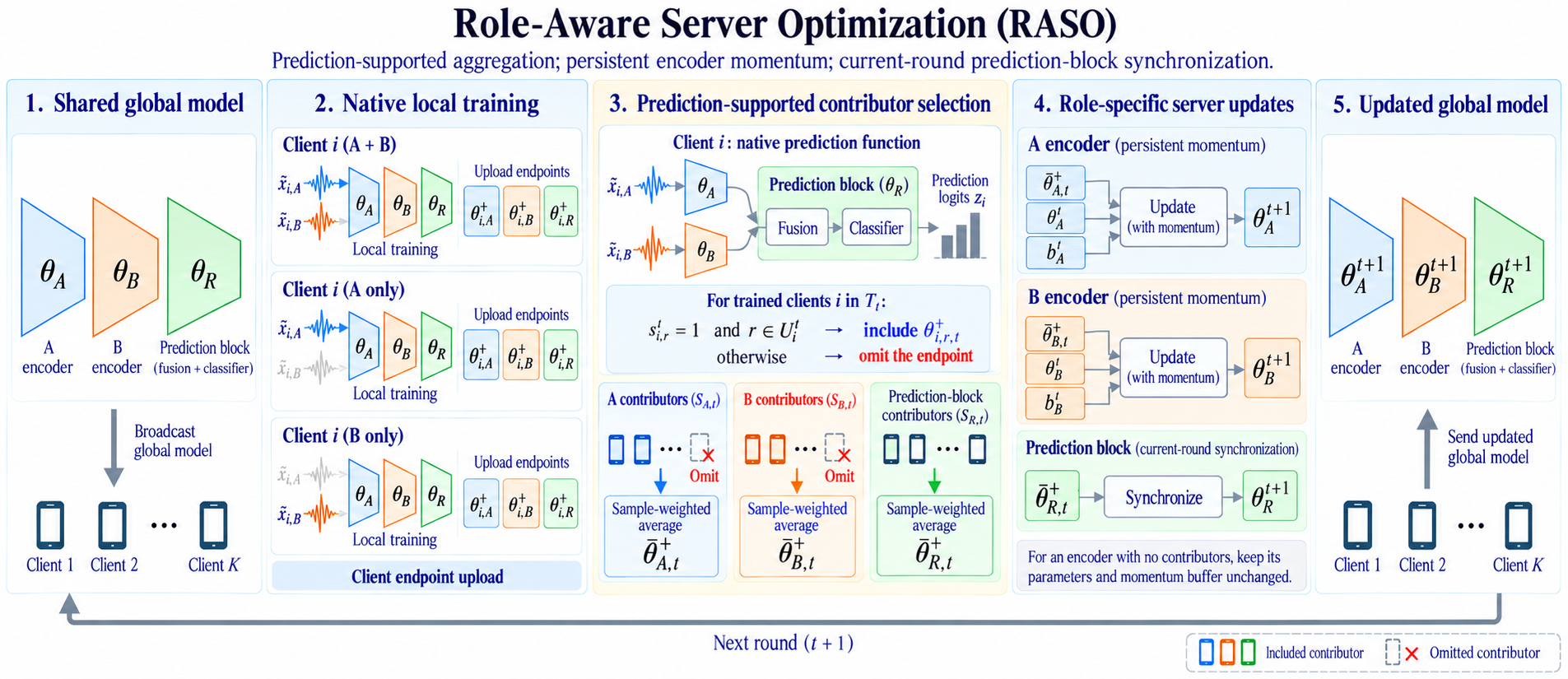}
\caption{
Overview of RASO. Prediction support and local optimization define
contributors $\mathcal S_{r,t}$, whose sample-weighted endpoints form
component aggregates. Encoder buffers accumulate endpoint residuals;
the prediction block follows its current aggregate. An encoder and
its buffer remain unchanged when its contributor set is empty.
}
\label{fig:raso_architecture}
\end{figure*}

RASO first selects and averages client parameters for each component,
then applies the corresponding server update
(Fig.~\ref{fig:raso_architecture}).

\subsection{Problem formulation}
\label{sec:problem}

Client $i$ holds a labeled dataset $\mathcal D_i$ of size $n_i$ and has
modality availability
$a_i^t=(a_{i,1}^t,\ldots,a_{i,M}^t)\in\{0,1\}^{M}$.
The model $\theta=(\theta_1,\ldots,\theta_M,\theta_R)$ comprises
modality encoders $\{\theta_m\}_{m=1}^{M}$ and a shared prediction
block $\theta_R$ containing the trainable fusion module and classifier.
Among the clients $\mathcal C_t$ selected at round $t$, those with at
least one available modality form
$\mathcal T_t=\{i\in\mathcal C_t:\sum_m a_{i,m}^t>0\}$.
Each trains from $\theta^t$ using its existing missing-input computation
and returns the resulting parameters, or endpoint, $\theta_{i,t}^{+}$.
If $\mathcal T_t=\varnothing$, the server model and optimizer state remain
unchanged.

\subsection{Prediction-supported aggregation}
\label{sec:support}

Let $p_i(\cdot;\theta,a_i^t)$ denote client $i$'s prediction function
under its missing-input representation and fusion rule. We call its
functional dependence on $\theta_r$~\cite{baydin2018automatic}
\emph{prediction support}. For $r\in\mathcal R=\{1,\ldots,M,R\}$,
\begin{equation}
s_{i,r}^t=
\mathbf 1\!\left\{
J_{\theta_r}p_i(\cdot;\theta,a_i^t)\not\equiv0
\right\}.
\label{eq:prediction_support}
\end{equation}
For fixed $a_i^t$, $s_{i,r}^t=1$ denotes dependence on block
$\theta_r$; realized minibatch gradients may still be zero.
Let $\mathcal U_i^t$ contain the blocks with trainable parameters included
in client $i$'s local optimizer. In RASO's evaluated implementations, concatenation retains
the absent encoder's zero-input response, whereas masked attention masks
its contribution. For modalities $A,B$,
\begin{equation}
(s_{i,A}^t,s_{i,B}^t)=
\begin{cases}
(1,1), & \text{concatenation},\\
(a_{i,A}^t,a_{i,B}^t), & \text{masked attention},
\end{cases}
\label{eq:native_support}
\end{equation}
with $s_{i,R}^t=1$ for every trained client. For these implementations,
encoder support is assigned directly by Eq.~\eqref{eq:native_support}.
Combining these indicators with local optimizer membership gives
\begin{equation}
\mathcal S_{r,t}
=
\{i\in\mathcal T_t:s_{i,r}^t=1,\ r\in\mathcal U_i^t\}.
\label{eq:contributor_set}
\end{equation}
For $\mathcal S_{r,t}\neq\varnothing$, RASO forms
\begin{equation}
\bar\theta_{r,t}^{+}
=
\frac{\sum_{i\in\mathcal S_{r,t}}
n_i\theta_{i,r,t}^{+}}
{\sum_{i\in\mathcal S_{r,t}}n_i}.
\label{eq:supported_endpoint}
\end{equation}
The selected clients are weighted by sample count, including those that
return unchanged parameters.

To quantify the discrepancy from availability-based selection, let
$\Delta_{i,m,t}=\theta_{i,m,t}^{+}-\theta_m^t$ and partition
$\mathcal S_{m,t}$ into modality-available and unavailable cohorts
$\mathcal A_{m,t}$ and $\mathcal O_{m,t}$. When both are nonempty,
let $\pi_A=\sum_{i\in\mathcal A_{m,t}}n_i$ and
$\pi_O=\sum_{i\in\mathcal O_{m,t}}n_i$ denote their sample masses,
with weighted mean displacements $\bar\Delta_A,\bar\Delta_O$. Then
\begin{equation}
\bar\Delta_S-\bar\Delta_A
=
\frac{\pi_O}{\pi_A+\pi_O}
(\bar\Delta_O-\bar\Delta_A),
\label{eq:contributor_discrepancy}
\end{equation}
where $\bar\Delta_S$ is the prediction-supported mean displacement.

\subsection{Role-specific server updates}
\label{sec:server_updates}

For encoder $m$, the selected clients determine the residual
$\theta_m^t-\bar\theta_{m,t}^{+}=-\bar\Delta_S$.
RASO accumulates this residual with momentum when
$\mathcal S_{m,t}\neq\varnothing$:
\begin{equation}
\begin{aligned}
b_m^{t+1}
&=\mu b_m^t+\theta_m^t-\bar\theta_{m,t}^{+},\\
\theta_m^{t+1}
&=\theta_m^t-\frac{\eta_s}{\eta_\ell}b_m^{t+1},
\end{aligned}
\qquad m=1,\ldots,M,
\label{eq:encoder_state}
\end{equation}
Here, $b_m^0=0$, $\mu$ is server momentum, and $\eta_\ell,\eta_s$ are
local and server step sizes. Empty contributor sets leave parameters and
buffers unchanged. A residual retains weight $\mu^k$ after $k$ subsequent
updates with contributors ($\mu^{10}\approx0.35$ at $\mu=0.9$).

The shared prediction block is synchronized to its current aggregate:
\begin{equation}
\theta_R^{t+1}=\bar\theta_{R,t}^{+},
\qquad b_R^{t+1}=0.
\label{eq:prediction_state}
\end{equation}
For fixed $\eta_\ell$, the scaled endpoint residual
$q_m^t=(\theta_m^t-\bar\theta_{m,t}^{+})/\eta_\ell$ and buffer
$v_m^t=b_m^t/\eta_\ell$ give the equivalent parameterization
$v_m^{t+1}=\mu v_m^t+q_m^t$ and
$\theta_m^{t+1}=\theta_m^t-\eta_s v_m^{t+1}$.
RASO stores one encoder-sized momentum buffer per modality, totaling
$\sum_{m=1}^{M}\dim(\theta_m)$ additional server-side scalars.

\section{Experiments}
\label{sec:experiments}

We evaluate UCI-HAR~\cite{anguita2013ucihar}, KU-HAR~\cite{sikder2021kuhar},
CREMA-D~\cite{cao2014cremad}, and CrisisMMD~\cite{alam2018crisismmd} with
two-modality concatenation and masked-attention models. Scores are
macro-F1, except CREMA-D UAR, in percent. FedMultimodal
~\cite{feng2023fedmultimodal} provides matched backbones, preprocessing,
partitions, modality assignments, and client schedules; methods retain
their local objectives, missing-input handling, and hyperparameters.
Training uses 10\% participation, one local epoch, batch size 16, local
learning rate $0.05$, local momentum $0.9$, and weight decay $10^{-5}$.
RASO and the persistent-encoder controls use $\eta_s=0.025$ and $\mu=0.9$. Terminal evaluation uses round 200; fixed-rate experiments use $q=0.5$. Fully missing clients are
skipped without replacement. TEST is excluded from hyperparameter and
checkpoint selection.

Terminal scores average three runs, except the single-run matched-rate sweep
(Fig.~\ref{fig:sensitivity}). Contributor diagnostics use concatenation;
support and availability coincide under masked attention. The benchmark
policy sweep and fixed-model tests include both fusions. The concatenation
factorial fixes fold 1 and reports global TEST scores with sample standard
deviations. Fixed-model tests reuse the original seed/fold models. Both
studies use frozen diagnostic TEST assignments shared across methods and
distinct from the main comparison. Replay contrasts average rounds
$25,50,\ldots,200$ within each of three runs, then across runs.

\subsection{Mechanism analysis}
\label{sec:observations}

\begin{table}[!t]
\centering
\fontsize{9}{10.5}\selectfont
\setlength{\tabcolsep}{3.0pt}
\renewcommand{\arraystretch}{1.03}
\caption{
Joint policies under masked attention with supported contributors.
Current encoders install the aggregate; persistent encoders use
Eq.~\eqref{eq:encoder_state} at gain $0.5$. Prediction blocks use
Eq.~\eqref{eq:prediction_controls} at $\lambda_R=1$ and $\rho_R=0/0.9$
(Current/Persistent). UCI/KU: macro-F1; CR: CREMA-D UAR (\%).
}
\label{tab:raso_lifecycle}

\begin{tabular}{@{}llrrr@{}}
\toprule
Encoders & Prediction block & UCI & KU & CR \\
\midrule
Current    & Current
           & 55.00 & 39.23 & 42.46 \\

Current    & Persistent
           & 56.17 & 47.82 & 45.77 \\

\textbf{Persistent}
           & \textbf{Current}
           & \textbf{63.33}
           & \textbf{48.30}
           & \textbf{48.52} \\

Persistent & Persistent
           & 5.31 & 32.47 & 42.27 \\
\bottomrule
\end{tabular}
\end{table}

\noindent\textit{Component participation and gradient direction.}
Figure~\ref{fig:gradient_observation} compares complete and single-missing
CE gradients from FedAvg and FedOpt at eight checkpoints over rounds
25--200. Each pair uses identical parameters and a 16-example training
minibatch. Under both fusions, the prediction block remains active, but its
gradient direction changes: cosines 0.433--0.756 ($41^\circ$--$64^\circ$).
Under concatenation, the absent encoder's gradient share
$r_{\rm abs}=\|g_{\rm abs}\|_2/(\|g_{\rm abs}\|_2+
\|g_{\rm present}\|_2)$ averages 0.206 on KU-HAR and 0.148 on CREMA-D;
$g_{\rm abs},g_{\rm present}$ are absent- and available-encoder CE gradients.
Its missing- and complete-input gradients have cosines 0.027--0.183
($79^\circ$--$88^\circ$). Masked attention gives zero absent-encoder
CE gradients while retaining the prediction-block redirection.

\noindent\textit{Joint encoder--prediction-block policies.}
Table~\ref{tab:raso_lifecycle} compares server-update policies under masked
attention with the local-training protocol and contributor rule fixed.
Prediction-block momentum raises scores with current encoder aggregation
but lowers them with persistent encoder updates on all three datasets.

\label{sec:support_discrepancy}
\noindent\textit{Contributor-selection diagnostics.}

For encoder--round pairs containing both availability cohorts,
we measure omitted sample-mass fraction $\gamma_{m,t}$, cohort
displacement cosine $c_{m,t}$, and relative aggregate change $d_{m,t}$:
\[
\begin{aligned}
\gamma_{m,t}&=\frac{\pi_O}{\pi_A+\pi_O},\qquad
c_{m,t}=\cos(\bar\Delta_A,\bar\Delta_O),\\
d_{m,t}&=\frac{\|\bar\Delta_S-\bar\Delta_A\|_2}
                  {\|\bar\Delta_S\|_2}.
\end{aligned}
\]
Table~\ref{tab:support_discrepancy}(a) shows that availability
filtering omits approximately one-third of the supported sample mass.
Median cohort cosine is near zero, and median relative aggregate
change is approximately 60\% on both datasets.

\begin{table}[!t]
\centering
\fontsize{9}{10.5}\selectfont
\setlength{\tabcolsep}{1.5pt}
\renewcommand{\arraystretch}{1.03}
\caption{
Concatenation diagnostics: (a) omitted mass $\gamma$ and median $c,d$
for encoder--round pairs with both cohorts; (b) mean signed DEV
cross-entropy contrasts ($10^{-3}$).
}
\label{tab:support_discrepancy}

\begin{tabular*}{\columnwidth}{@{\extracolsep{\fill}}lrrrrrrr@{}}
\toprule
& \multicolumn{3}{c}{\textit{(a) Aggregates}}
& \multicolumn{4}{c}{\textit{(b) Replay}}\\
\cmidrule(lr){2-4}\cmidrule(lr){5-8}
Dataset & $\gamma$ & $c$ & $d$
        & $B_C$ & $B_H$ & $I$ & $\Gamma$\\
\midrule
KU-HAR  & 33.3\% & $\approx 0$ & 63.1\%
        & $-4.385$ & $-7.645$ & $+3.260$ & $-5.408$\\
CREMA-D & 35.8\% & $\approx 0$ & 59.1\%
        & $-0.142$ & $-0.801$ & $+0.659$ & $-1.088$\\
\bottomrule
\end{tabular*}
\end{table}

\begin{table*}[!t]
\centering
\fontsize{9}{10.5}\selectfont
\setlength{\tabcolsep}{4pt}
\renewcommand{\arraystretch}{1.00}
\caption{
Main test-set comparison (\%). Macro-F1 is used except for CREMA-D
(UAR). Best mean in each column is bold.
}
\label{tab:main_results}

\begin{tabular*}{\textwidth}{@{\extracolsep{\fill}}lrrrrrrrr@{}}
\toprule
& \multicolumn{4}{c}{Masked attention}
& \multicolumn{4}{c}{Concatenation}\\
\cmidrule(lr){2-5}\cmidrule(lr){6-9}
Method & UCI-HAR & KU-HAR & CREMA-D & CrisisMMD
       & UCI-HAR & KU-HAR & CREMA-D & CrisisMMD\\
\midrule

FedAvg~\cite{mcmahan2017fedavg}
& 49.33 & 41.46 & 41.64 & 9.10
& 50.65 & 33.68 & 44.41 & 9.40\\

FedOpt~\cite{reddi2021fedopt}
& 43.72 & 32.54 & 43.49 & \textbf{11.00}
& 55.61 & 42.07 & 44.98 & 10.70\\

FedProx~\cite{li2020federated}
& 42.72 & 34.54 & 40.49 & 10.03
& 52.21 & 40.09 & 42.80 & 10.20\\

PmcmFL~\cite{bao2026pmcmfl}
& 49.45 & 43.29 & 41.80 & 8.70
& 52.14 & 38.91 & 40.98 & 9.10\\

MFCPL~\cite{le2025mfcpl}
& 58.66 & 35.66 & 40.21 & 10.10
& 53.72 & 31.32 & 40.16 & 9.80\\

FedMobile~\cite{liu2025fedmobile}
& 62.96 & 45.66 & 43.19 & 10.23
& \textbf{59.92} & \textbf{50.32} & 46.62 & \textbf{11.28}\\

BMSFed~\cite{fan2024overcome}
& 38.63 & 42.08 & 41.34 & 8.80
& 30.23 & 46.50 & 41.79 & 9.20\\

BLOSSOM~\cite{pranav2026blossom}
& 48.32 & 39.41 & 41.31 & 8.50
& 38.05 & 34.55 & 41.10 & 8.90\\

\midrule
\textbf{RASO}
& \textbf{63.33} & \textbf{48.30} & \textbf{48.52} & 10.50
& 59.41 & 49.13 & \textbf{48.20} & 10.80\\
\bottomrule
\end{tabular*}
\end{table*}

Matched-step replay compares immediate development (DEV) losses from
common model and optimizer states, client endpoints, and inputs. Encoder
step norms are matched component-wise across contributor rules;
prediction-block step norms are matched across current and historical directions.
Let $\ell_{a,b}$ be DEV cross-entropy, with $a\in\{S,A,0\}$ denoting
support-based, availability-based, or unchanged encoders, and $b\in\{C,H\}$
current or historical prediction-block directions:
\[
\begin{aligned}
B_b &= \ell_{A,b}-\ell_{S,b},\qquad I=B_C-B_H,\\
\Gamma &= (\ell_{S,H}-\ell_{S,C})-(\ell_{0,H}-\ell_{0,C}).
\end{aligned}
\]
In Table~\ref{tab:support_discrepancy}(b), availability-based candidates
have lower immediate DEV loss under both directions ($B_C,B_H<0$),
with a larger advantage under history ($I>0$). Negative mean $\Gamma$
indicates a decrease in the signed historical-minus-current loss
difference after the support-based encoder update.

\subsection{Performance and ablations}

Table~\ref{tab:main_results} shows RASO leading UCI-HAR, KU-HAR, and
CREMA-D under masked attention; FedOpt leads CrisisMMD. Under concatenation,
RASO leads CREMA-D and exceeds FedOpt on all four datasets, while FedMobile
leads the other three.

\noindent\textit{Server-update ablations.}
\label{sec:ablations}

We compare encoder contributor rules and prediction-block update policies
in matched 200-round training from initialization. The encoder update rule
and hyperparameters in Eq.~\eqref{eq:encoder_state} remain fixed.
For buffer $u_R$, define
$d_R^t=\theta_R^t-\bar\theta_{R,t}^{+}$ and use
\begin{equation}
\begin{aligned}
u_R^{t+1} &= \rho_R u_R^t+d_R^t,\\
\theta_R^{t+1} &= \theta_R^t-\lambda_R u_R^{t+1}.
\end{aligned}
\label{eq:prediction_controls}
\end{equation}
Here, $u_R^0=0$, $\rho_R$ controls history retention, and
$\lambda_R$ scales the buffered update. Current synchronization
(RASO) uses $(\rho_R,\lambda_R)=(0,1)$ in all main comparisons.
Normalized momentum uses $(0.9,0.1)$, with unit steady-state gain
$\lambda_R/(1-\rho_R)=1$ for a constant residual.

\begin{table}[!t]
\centering
\fontsize{9}{10.5}\selectfont
\setlength{\tabcolsep}{2pt}
\renewcommand{\arraystretch}{1.02}
\caption{
Server ablations: (a) benchmark; (b) fixed-fold concatenation
(mean $\pm$ sample SD). Support/Avail.: prediction support/availability;
Current/Norm.: $(\rho_R,\lambda_R)=(0,1)/(0.9,0.1)$,
both with unit steady-state gain for a constant residual.
KU: macro-F1; CR: UAR (\%). Bold marks RASO.
}
\label{tab:raso_attribution}

\begin{tabular*}{\columnwidth}{@{\extracolsep{\fill}}lccrrrr@{}}
\toprule
\multicolumn{7}{c}{\textit{(a) Benchmark policy sweep}}\\
&&& \multicolumn{2}{c}{Masked attention}
  & \multicolumn{2}{c}{Concatenation}\\
\cmidrule(lr){4-5}\cmidrule(lr){6-7}
Rule & $\rho_R$ & $\lambda_R$ & KU & CR & KU & CR\\
\midrule
\textbf{Support} & \textbf{0} & \textbf{1}
 & \textbf{48.30} & \textbf{48.52}
 & \textbf{49.13} & \textbf{48.20}\\
Avail. & 0 & 1
 & 48.30 & 48.52 & 41.30 & 43.00\\
Support & 0.3 & 1
 & 48.75 & 45.67 & 46.30 & 43.59\\
Support & 0.6 & 1
 & 44.68 & 38.93 & 43.37 & 43.56\\
Support & 0.9 & 1
 & 32.47 & 42.27 & 39.07 & 42.61\\
Avail. & 0.9 & 1
 & 32.47 & 42.27 & 37.15 & 38.51\\
\midrule
Support & 0 & 0.5
 & 40.35 & 36.62 & 43.56 & 33.70\\
Support & 0.9 & 0.5
 & 15.14 & 37.96 & 33.58 & 39.58\\
Avail. & 0.9 & 0.5
 & 15.14 & 37.96 & 42.87 & 42.76\\
\midrule
Support & 0.9 & 0.1
 & 38.46 & 45.59 & 44.71 & 44.45\\
\midrule
\end{tabular*}

\par\vspace{2pt}
\setlength{\tabcolsep}{1pt}
\begin{tabular*}{\columnwidth}{@{\extracolsep{\fill}}lrrrr@{}}
\multicolumn{5}{c}{\textit{(b) Fixed-fold factorial: concatenation}}\\
& \multicolumn{2}{c}{Availability}
& \multicolumn{2}{c}{Support}\\
\cmidrule(lr){2-3}\cmidrule(lr){4-5}
Dataset & Current & Norm. & Current & Norm.\\
\midrule
KU & $36.15\pm7.82$ & $39.64\pm2.76$
   & $\mathbf{58.18}\pm1.42$ & $49.18\pm3.23$\\
CR & $52.91\pm0.89$ & $48.59\pm0.68$
   & $\mathbf{54.77}\pm0.41$ & $51.36\pm1.02$\\
\bottomrule
\end{tabular*}

\par\medskip
\setlength{\tabcolsep}{2.5pt}
\caption{
Fixed-model TEST scores averaged over rates and runs (\%).
UCI/KU: macro-F1; CR: UAR. Bold marks column maxima.
}
\label{tab:missingness_shift}
\setlength{\tabcolsep}{1.3pt}
\begin{tabular*}{\columnwidth}{@{\extracolsep{\fill}}lrrrrrr@{}}
\toprule
& \multicolumn{3}{c}{Masked attention}
& \multicolumn{3}{c}{Concatenation}\\
\cmidrule(lr){2-4}\cmidrule(lr){5-7}
Method & UCI & KU & CR & UCI & KU & CR\\
\midrule
FedAvg & 52.19 & 43.94 & 47.22 & 55.13 & 35.95 & 50.07\\
FedOpt & 45.90 & 35.96 & 47.05 & 59.64 & 45.72 & 48.28\\
PmcmFL & 53.75 & 47.85 & 43.65 & 56.33 & 44.96 & 45.91\\
\textbf{RASO} & \textbf{63.90} & \textbf{51.70} & \textbf{53.01}
 & \textbf{64.13} & \textbf{52.51} & \textbf{51.32}\\
\bottomrule
\end{tabular*}
\end{table}

\begin{figure}[!t]
\centering
\includegraphics[width=\columnwidth]{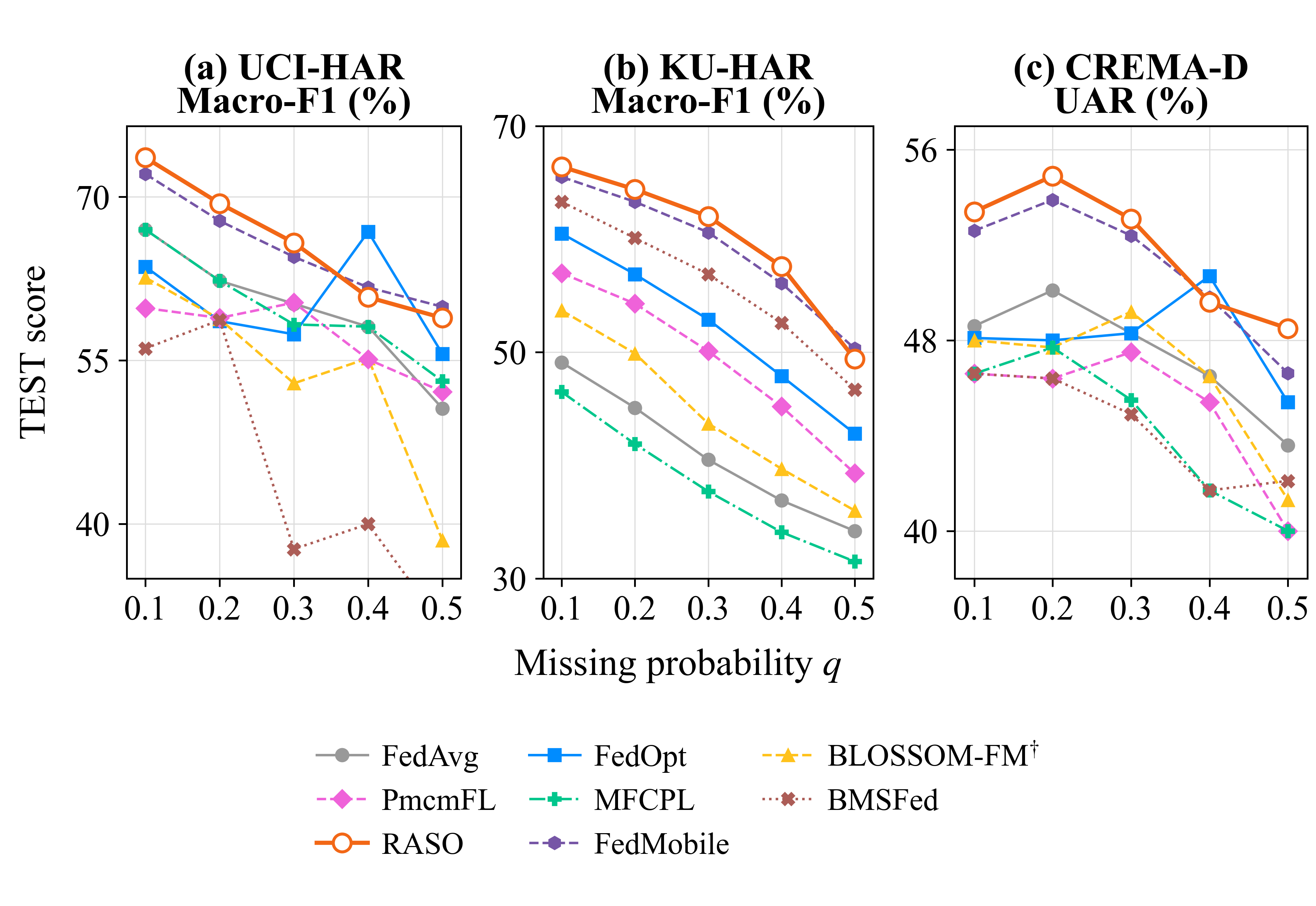}
\caption{
Single-run concatenation sweep, with a separately trained model at each
matched training/test missingness rate. UCI-HAR/KU-HAR: macro-F1;
CREMA-D: UAR (\%).
}
\label{fig:sensitivity}
\end{figure}

\begin{samepage}
In the fixed-fold concatenation factorial
(Table~\ref{tab:raso_attribution}(b)), support-based aggregation yields
higher scores under both prediction-block policies, and RASO ranks first
on both datasets. The support advantage is larger with current
synchronization on KU-HAR and normalized momentum on CREMA-D.
\par
\end{samepage}

The benchmark sweep examines prediction-block momentum and gain
(Table~\ref{tab:raso_attribution}(a)). At full gain, support peaks at
$\rho_R=0.3$ for masked-attention KU-HAR and $0$ otherwise.
Under concatenation, support leads the matched full-gain comparisons
($\rho_R=0,0.9$), whereas availability leads at $(0.9,0.5)$.
With supported contributors, increasing $\rho_R$ from $0$ to $0.9$
at half gain lowers KU-HAR and raises CREMA-D under both fusions;
$(0.9,0.1)$ improves all four means over $(0.9,1)$.

\subsection{Missingness evaluation}
\label{sec:missingness_shift}

Figure~\ref{fig:sensitivity} compares separately trained concatenation
models at matched training/test rates $q\in\{0.1,0.2,\ldots,0.5\}$.
Methods share partitions, modality assignments, schedules, and
participating clients within each rate; skipping fully missing clients
can change the population across rates.

Table~\ref{tab:missingness_shift} tests unchanged $q=0.5$ models at
these rates. Among the four methods, RASO has the highest rate-averaged
scores on all three datasets under both fusions and leads all but one
per-rate comparison: concatenation CREMA-D at $q_{\rm test}=0.1$ favors
FedAvg (53.99 versus 53.55).

\section{Conclusion}
RASO couples prediction-supported aggregation with persistent encoder
momentum and current prediction-block synchronization for MFL with
missing modalities. Full-training comparisons support this policy, while
sweeps and replay show interactions between contributors and server updates.

\bibliographystyle{IEEEbib}
\bibliography{strings,refs}

@inproceedings{mcmahan2017fedavg,
  author    = {McMahan, Brendan and Moore, Eider and Ramage, Daniel and Hampson, Seth and Arcas, Blaise Aguera y},
  title     = {Communication-Efficient Learning of Deep Networks from Decentralized Data},
  booktitle = {Proceedings of the 20th International Conference on Artificial Intelligence and Statistics},
  series    = {Proceedings of Machine Learning Research},
  volume    = {54},
  pages     = {1273--1282},
  year      = {2017},
  publisher = {PMLR},
  url       = {https://proceedings.mlr.press/v54/mcmahan17a.html}
}

@inproceedings{feng2023fedmultimodal,
  author    = {Feng, Tiantian and Bose, Digbalay and Zhang, Tuo and Hebbar, Rajat and Ramakrishna, Anil and Gupta, Rahul and Zhang, Mi and Avestimehr, Salman and Narayanan, Shrikanth},
  title     = {{FedMultimodal}: A Benchmark for Multimodal Federated Learning},
  booktitle = {Proceedings of the 29th ACM SIGKDD Conference on Knowledge Discovery and Data Mining},
  pages     = {4035--4045},
  year      = {2023},
  publisher = {ACM},
  doi       = {10.1145/3580305.3599825},
  url       = {https://doi.org/10.1145/3580305.3599825}
}

@article{bao2026pmcmfl,
  author  = {Bao, Guangyin and Zhang, Qi and Miao, Duoqian and Gong, Zixuan and Chen, Chaochao and Hu, Liang and Cao, Longbing},
  title   = {Incomplete Multimodal Federated Learning via Masking and Contrasting Prototypes},
  journal = {IEEE Transactions on Neural Networks and Learning Systems},
  year    = {2026},
  doi     = {10.1109/TNNLS.2026.3658522},
  url     = {https://doi.org/10.1109/TNNLS.2026.3658522},
  note    = {Early access / ahead of print}
}

@article{le2025mfcpl,
  author  = {Le, Huy Q. and Thwal, Chu Myaet and Qiao, Yu and Tun, Ye Lin and Nguyen, Minh N. H. and Huh, Eui Nam and Hong, Choong Seon},
  title   = {Cross-modal prototype based multimodal federated learning under severely missing modality},
  journal = {Information Fusion},
  volume  = {122},
  pages   = {103219},
  year    = {2025},
  doi     = {10.1016/j.inffus.2025.103219},
  url     = {https://doi.org/10.1016/j.inffus.2025.103219}
}

@inproceedings{fan2024overcome,
  author    = {Fan, Yunfeng and Xu, Wenchao and Wang, Haozhao and Huo, Fushuo and Chen, Jinyu and Guo, Song},
  title     = {Overcome Modal Bias in Multi-modal Federated Learning via Balanced Modality Selection},
  booktitle = {Computer Vision -- ECCV 2024},
  series    = {Lecture Notes in Computer Science},
  volume    = {15139},
  pages     = {178--195},
  year      = {2024},
  publisher = {Springer},
  doi       = {10.1007/978-3-031-73004-7_11},
  url       = {https://doi.org/10.1007/978-3-031-73004-7_11}
}

@inproceedings{chen2022fedmsplit,
  author    = {Chen, Jiayi and Zhang, Aidong},
  title     = {{FedMSplit}: Correlation-Adaptive Federated Multi-Task Learning across Multimodal Split Networks},
  booktitle = {Proceedings of the 28th ACM SIGKDD Conference on Knowledge Discovery and Data Mining},
  pages     = {87--96},
  year      = {2022},
  publisher = {ACM},
  doi       = {10.1145/3534678.3539384},
  url       = {https://doi.org/10.1145/3534678.3539384}
}

@article{pranav2026blossom,
  author  = {{Pranav M R} and Chandwani, Jayant and Abdelmoniem, Ahmed M. and Paul, Arnab K.},
  title   = {{BLOSSOM}: Block-wise Federated Learning Over Shared and Sparse Observed Modalities},
  journal = {arXiv preprint arXiv:2603.27552},
  year    = {2026},
  url     = {https://arxiv.org/abs/2603.27552},
  note    = {Accepted/presented at IJCNN 2026}
}

@inproceedings{alam2022fedrolex,
  author    = {Alam, Samiul and Liu, Luyang and Yan, Ming and Zhang, Mi},
  title     = {{FedRolex}: Model-Heterogeneous Federated Learning with Rolling Sub-Model Extraction},
  booktitle = {Advances in Neural Information Processing Systems},
  volume    = {35},
  pages     = {29677--29690},
  year      = {2022},
  doi       = {10.52202/068431-2152},
  url       = {https://proceedings.neurips.cc/paper_files/paper/2022/hash/bf5311df07f3efce97471921e6d2f159-Abstract-Conference.html}
}

@inproceedings{reddi2021fedopt,
  author    = {Reddi, Sashank J. and Charles, Zachary and Zaheer, Manzil and Garrett, Zachary and Rush, Keith and Kone{\v{c}}n{\'y}, Jakub and Kumar, Sanjiv and McMahan, H. Brendan},
  title     = {Adaptive Federated Optimization},
  booktitle = {International Conference on Learning Representations},
  year      = {2021},
  url       = {https://openreview.net/forum?id=LkFG3lB13U5}
}

@inproceedings{li2023fedpvr,
  author    = {Li, Bo and Schmidt, Mikkel N. and Alstr{\o}m, Tommy S. and Stich, Sebastian U.},
  title     = {On the Effectiveness of Partial Variance Reduction in Federated Learning With Heterogeneous Data},
  booktitle = {Proceedings of the IEEE/CVF Conference on Computer Vision and Pattern Recognition},
  pages     = {3964--3973},
  year      = {2023},
  doi       = {10.1109/CVPR52729.2023.00386},
  url       = {https://openaccess.thecvf.com/content/CVPR2023/html/Li_On_the_Effectiveness_of_Partial_Variance_Reduction_in_Federated_Learning_CVPR_2023_paper.html}
}

@inproceedings{anguita2013ucihar,
  author    = {Anguita, Davide and Ghio, Alessandro and Oneto, Luca and Parra, Xavier and Reyes-Ortiz, Jorge L.},
  title     = {A Public Domain Dataset for Human Activity Recognition Using Smartphones},
  booktitle = {Proceedings of the European Symposium on Artificial Neural Networks, Computational Intelligence and Machine Learning (ESANN)},
  pages     = {437--442},
  year      = {2013},
  url       = {https://www.esann.org/sites/default/files/proceedings/legacy/es2013-84.pdf}
}

@article{sikder2021kuhar,
  author  = {Sikder, Niloy and Nahid, Abdullah-Al},
  title   = {{KU-HAR}: An Open Dataset for Heterogeneous Human Activity Recognition},
  journal = {Pattern Recognition Letters},
  volume  = {146},
  pages   = {46--54},
  year    = {2021},
  doi     = {10.1016/j.patrec.2021.02.024},
  url     = {https://doi.org/10.1016/j.patrec.2021.02.024}
}

@article{cao2014cremad,
  author  = {Cao, Houwei and Cooper, David G. and Keutmann, Michael K. and Gur, Ruben C. and Nenkova, Ani and Verma, Ragini},
  title   = {{CREMA-D}: Crowd-Sourced Emotional Multimodal Actors Dataset},
  journal = {IEEE Transactions on Affective Computing},
  volume  = {5},
  number  = {4},
  pages   = {377--390},
  year    = {2014},
  doi     = {10.1109/TAFFC.2014.2336244},
  url     = {https://doi.org/10.1109/TAFFC.2014.2336244}
}

@article{alam2018crisismmd,
  author  = {Alam, Firoj and Ofli, Ferda and Imran, Muhammad},
  title   = {{CrisisMMD}: Multimodal Twitter Datasets from Natural Disasters},
  journal = {Proceedings of the International AAAI Conference on Web and Social Media},
  volume  = {12},
  number  = {1},
  year    = {2018},
  doi     = {10.1609/icwsm.v12i1.14983},
  url     = {https://ojs.aaai.org/index.php/ICWSM/article/view/14983}
}

@article{li2020federated,
  author  = {Li, Tian and Sahu, Anit Kumar and Zaheer, Manzil and Sanjabi, Maziar and Talwalkar, Ameet and Smith, Virginia},
  title   = {Federated Optimization in Heterogeneous Networks},
  journal = {Proceedings of Machine Learning and Systems},
  volume  = {2},
  pages   = {429--450},
  year    = {2020},
  url     = {https://proceedings.mlsys.org/paper_files/paper/2020/hash/1f5fe83998a09396ebe6477d9475ba0c-Abstract.html}
}

@inproceedings{10.1007/978-981-92-3520-9_47,
  author    = {Zhao, Qiheng and Feng, Siwei},
  title     = {{FedGMM}: Robust Temporal Multimodal Federated Learning Under Missing Modalities},
  booktitle = {Advanced Intelligent Computing Technology and Applications},
  series    = {Lecture Notes in Computer Science},
  volume    = {16672},
  pages     = {592--603},
  year      = {2027},
  publisher = {Springer Nature Singapore},
  address   = {Singapore},
  doi       = {10.1007/978-981-92-3520-9_47},
  url       = {https://doi.org/10.1007/978-981-92-3520-9_47},
  isbn      = {978-981-92-3520-9},
  note      = {ICIC 2026; first online 19 July 2026}
}

@article{yuan2026communication,
  author    = {Yuan, Liangqi and Han, Dong-Jun and Wang, Su and Upadhyay, Devesh and Brinton, Christopher G.},
  title     = {Communication-Efficient Multimodal Federated Learning: Joint Modality and Client Selection},
  journal   = {IEEE Transactions on Mobile Computing},
  volume    = {25},
  number    = {8},
  pages     = {12780--12796},
  year      = {2026},
  publisher = {IEEE},
  doi       = {10.1109/TMC.2026.3673697},
  url       = {https://doi.org/10.1109/TMC.2026.3673697}
}

@inproceedings{yuan2024fedmfs,
  title={FedMFS: Federated multimodal fusion learning with selective modality communication},
  author={Yuan, Liangqi and Han, Dong-Jun and Chellapandi, Vishnu Pandi and Zak, Stanislaw H and Brinton, Christopher G},
  booktitle={ICC 2024-IEEE International Conference on Communications},
  pages={287--292},
  year={2024},
  organization={IEEE}
}

@inproceedings{liu2025fedmobile,
  title={Fedmobile: Enabling knowledge contribution-aware multi-modal federated learning with incomplete modalities},
  author={Liu, Yi and Wang, Cong and Yuan, Xingliang},
  booktitle={Proceedings of the ACM on Web Conference 2025},
  pages={2775--2786},
  year={2025}
}

@article{baydin2018automatic,
  title={Automatic differentiation in machine learning: a survey},
  author={Baydin, Atilim Gunes and Pearlmutter, Barak A and Radul, Alexey Andreyevich and Siskind, Jeffrey Mark},
  journal={Journal of machine learning research},
  volume={18},
  number={153},
  pages={1--43},
  year={2018}
}

@inproceedings{che2024leveraging,
  title={Leveraging foundation models for multi-modal federated learning with incomplete modality},
  author={Che, Liwei and Wang, Jiaqi and Liu, Xinyue and Ma, Fenglong},
  booktitle={Joint European Conference on Machine Learning and Knowledge Discovery in Databases},
  pages={401--417},
  year={2024},
  organization={Springer}
}

@inproceedings{nguyen2024fedmac,
  title={Fedmac: Tackling partial-modality missing in federated learning with cross-modal aggregation and contrastive regularization},
  author={Nguyen, Manh Duong and Nguyen, Trung Thanh and Pham, Huy Hieu and Hoang, Trong Nghia and Le Nguyen, Phi and Huynh, Thanh Trung},
  booktitle={2024 22nd International Symposium on Network Computing and Applications (NCA)},
  pages={278--285},
  year={2024},
  organization={IEEE}
}

@article{nguyen2026learning,
  title={Learning reconfigurable representations for multimodal federated learning with missing data},
  author={Nguyen, Duong and Hoang, Nghia and Huynh, Thanh Trung and Nguyen, Quoc Viet Hung and Nguyen, Phi Le},
  journal={Advances in Neural Information Processing Systems},
  volume={38},
  pages={28808--28845},
  year={2026}
}

\end{document}